# A Novel Sample-efficient Deep Reinforcement Learning with Episodic Policy Transfer for PID-Based Control in Robotic Catheter System

Olatunji Mumini OMISORE, Toluwanimi Oluwadara AKINYEMI, Wenke DUAN, Wenjing DU and Lei WANG

*Abstract*— Robotic catheterization is typically used for percutaneous coronary intervention procedures nowadays and it involves steering flexible endovascular tools to open up occlusion in the coronaries. In this study, a sample-efficient deep reinforcement learning with episodic policy transfer is, for the first time, used for motion control during robotic catheterization with fully adaptive PID tuning strategy. The reinforcement model aids the agent to continuously learn from its interactions in its environment and adaptively tune PID control gains for axial navigation of endovascular tool. The model was validated for axial motion control of a robotic system designed for intravascular catheterization. Simulation and experimental trials were done to validate the application of the model, and results obtained shows it could self-tune PID gains appropriately for motion control of a robotic catheter system. Performance comparison with conventional methods in average of 10 trials shows the agent tunes the gain better with error of 0.003 mm. Thus, the proposed model would offer more stable set-point motion control robotic catheterization.

## I. INTRODUCTION

Cardiovascular diseases are mal-conditions of vessels that transport blood around the heart and these remained a leading cause of global deaths [1]. Percutaneous coronary intervention (PCI) has been used as a substitute to open-heart approach. It involves catheterization of endovascular tools from peripheral ports into cardiac regions for better care optimization, improved cosmesis and surgical outcome [2]. This flexible access surgical approach has recently gained significant procedural advances in cardiac diagnoses and therapies. With it, interventionists can directly access a beating heart via intravascularly. The endovascular tools are long thin flexible tubes (*catheter*) and wires (*guidewire*) that are passed into the heart via the vascular system. While the tools can perform many tasks inside the heart, their intravascular manipulation requires proper training for precise navigation and safe interaction with vessel and heart tissues, a current gap in the manual catheterization methods that is still widely adopted [3, 4]. In addition, PCI involves cannulation along tortuous vascular paths under x-ray getting surgeons exposed to radiation. It gets worse since last decade as increased prevalence of diseases like atherosclerosis and thrombosis requires interventionists appear often in radiation rooms [5].

This work was supported by National Key Research and Development program of China (#2019YFB1311700); the National Natural Science Foundation of China (#U1713219, #61950410618); Shenzhen Natural Science Foundation (#JCYJ20190812173205538); and CAS President's International Fellowship Initiative for postdoctoral studies.

O.M. Omisore, T. O. Akinyemi, W.K. Duan, W.J. Du, and L. Wang are with Research Center for Medical Robotics and MIS Devices, Shenzhen Institutes of Advanced Technology, Chinese Academy of Sciences, Shenzhen, China; {omisore, tolu, wk.duan, wj.du, wang.lei}@siat.ac.cn.

Recently, robotic catheter systems (RCSs) have been used to reduce the human sacrifice during percutaneous coronary intervention and the robot-assisted intervention has become an impeccable alternative to the conventional PCI method. Commercial robotic catheter systems (RCSs), such as CorPath 200 (Corindus Vascular Robotics), Sensei (Hansen Medical Inc), Niobe (Stereotaxis Inc.) and Amigo (Catheter Precision) are now used in some parts of the world for safe and effective catheterization. Chakravartti *et al.* [2] presented that robotic PCI provides diverse benefits that both surgical team (*e.g.* the interventionist are now free from radiation) and the patients. However, the benefits enumerated for patients are not significantly different from the iconic achievements in the era of conventional PCI. Patel *et al.* [2] compared a large cohort of robotic and conventional PCI procedures that were performed around the same time in a tertiary care center, and reported that dose area product was lesser during the robot-assisted PCIs. Similarly, the study shows a substantial reduction in the exposure received by both doctors and patient while the use of contrast agent and fluoroscopy time is *at par* in both. Despite the evidences available for safety and efficiency of these RCSs, robot-assisted cardiac interventions are only carried out in few clinical centres, world-wide [5].

The underactuation mechanism in RCSs backlashed the development of effective control systems for intravascular navigation of the endovascular tools during catheterization. Research intensity is high to develop motion and force based control methods that can aid effective operation of RCSs. However, precise positioning of the flexible tools is still negatively affected by hysteresis in current control strategies [7]. This leads to improper tool navigation which makes the blood vessel susceptible to fatal damages such as severing, puncture, or thrombosis. To avoid unnecessary vessel damage from imprecise positioning, motion control systems have been used to characterize the causal factors of hysteresis from tool motion and actuator points of view, while online compensation schemes have been developed to compensate the motion errors. For instance, Kesner & Howe [4] developed a position-based motion compensation system for robotic catheter navigation during intracardiac interventions. In similar studies, other authors have also developed model-based and model-free navigation control methods [7-9]. For instance, Feng *et al.* [10] developed a simple speed-based control for scaling master-slave motion of a RCS used for vascular interventions. While the control methods could govern navigation of flexible tools handled in the underactuated robots, the main idea centralized on dynamical effects such as friction and backlash outside the mechanical systems and their compensation for navigation.

To better posit this problem in control engineering, use of classical control methods in flexible access robot domain are evaluated. Conventionally, cascaded configuration of PID controllers has dominated the robotic control space. This involved tuning of different gains for smooth motion and force control and has been applied in of robotics kinematics, dynamics, and teleoperation control of systems. In a mater-slave robotic system designed by Ma *et al.* [11], a conventional model of PID was implemented on the master side for navigation of flexible endovascular tools via direct slave robot operation. Thus, the master device produced more stable motion over the operating range. Similarly, Sariyildiz *et al.* [12] developed a robust PID controller with velocity feed-back for motion control of RCS used in vascular intervention robots. These PID controllers operate on fixed gains and cannot provide the performance consistency in terms of error compensation needed for PCIs. Alternatively, adaptive PID controllers have been developed. Zhao *et al.* [13] reported a self-tuning fuzzy-PID control system for 2-degree of freedom (DoF) motion in an interventional robot. Guo *et al.* [14] cascaded fuzzy-PID units for motion control of guidewire with force feedback, and appropriate unit was selected according to navigation situations during procedures. In Guo *et al.* [15], fuzzy-PID controller was rather used for feedback-based force control validated with Matlab simulation of a vascular interventional robot. Shi *et al.* [16] applied a PID controller enhanced on neural network for remote motion navigation of a RCS and compared its performance with conventional PID control. Also, design variations in fuzzy-PID control systems have been examined in robotics domain. Recently, For instance, Muskinja & Riznar [17] established a non-model tuning strategy in a dual PD controller cascaded for motion control in a ball and beam system. Omisore *et al.* [18] also developed a fuzzy-PD for teleoperation control of two Omni haptic robots in master-slave setup. Similarly, a few studies have been recently found in RCSs. Yu *et al.* [19] developed a dual Fuzzy-PID for online parameter tuning and interference removal in a RCS. Overall, the classical and adaptive PID controllers provide online gain tuning to achieve better performance such as more precise motion control in RCSs.

Despite the progresses cited above, literature overview shows that the motivation for "adaptability" in PID-based motion control in flexible access surgical robots is yet to be achieved [6]. The current designs are not currently able to self-adapt in their application domain. For instance cases of communication and control issues that causes strong hysteresis [1, 4, 7-9] or changes that require kinematics and dynamics design alteration [17] would require retuning a classical PID or at least redefining new rules or data needed to extend tuning in the classical PIDs with intelligence. We thereby, in this study, hypothesize that reinforcement learning can aid PID control system with extended self-tuning and self-adaptability to achieve better motion control performance and improved behavior in the above raised cases and ultimately achieve a given intravascular catheterization task (goal) within partially unknown and uncertain environment conditions. In this study, a sample-efficient deep reinforcement learning with episodic policy transfer is, for the first time, used for motion control of a RCS with fully adaptive PID tuning strategy. We limit the task to bi-axial catheterization of our newly designed multi-DoF RCS [19] in a simulated surgical environment. Remainder of this paper is organized such that the details of the RCS and deep reinforcement learning PID (DRL-PID) developed for intravascular catheterization are presented in Section II; implementation details of the deep reinforcement learning and analysis are presented in Section III; validation and evaluation studies carried out for performance analysis are presented in Section IV; finally, the conclusion of the study and future works are discussed in Section V.

## II. SYSTEM DESIGN AND MODELING

During typical procedures, flexible endovascular tools such as catheter and guidewire are robotically steered to cannulate arteries and open up occlusions in the coronary. In this study, this is considered as a two-part engineering problem of designing a multi-axial RCS and developing a self-adaptive motion control system for navigation of the flexible tools during intravascular cardiac catheterizations.

### A. Design of Robotic Catheter System

Design of our RCS is based on iterative prototyping from the first and second early generations presented in our studies [1, 19-20]. The current generation is an isomorphic master-slave RCS design proposed towards task specific autonomy during intravascular catheterization. The slave-side device, current focus in this study, is a robotic 4-DoF mechanism designed for intravascular interventions in this study is displayed in Fig. 1. The mechanism utilizes two distinct motions namely axial translation (*push* or *pull*) and axial rotation commands received from the master-side device (details are in [20]). The slave-side device is small dimensioned (57×22×16 cm), thus the drive system would little-to-no interference that distorts the actuation. Unlike the earlier generations, the current prototype has in-built sensing units for position and force feedback, and motion commands received from the master interface are sampled over a TCP/IP network for a directly scaled motion control system. Currently, the network protocol only induces a distributed minimal delay ($\cong 0.1$ sec), while different movements of axial translation, axial rotation, and their hybrid are implemented to mimic natural operational methods followed by interventionists in the interventional rooms. The system is self-powered to avoid electrical interference and exposure of patients to shock in the operational room. Thus, the power distribution methods discussed in the second generation [1] is adopted. Details of other components and their functionality are skipped as they are beyond the scope of this paper.

### B. Reinforcement PID Modeling for RCS Motion Control

The robotic mechanism in Fig.1 is capable of both axial translation and rotation during a typical intravascular catheterization procedure. Since both motion axes are independent, each can be characterized uniquely over its tangential direction. Motion of robot's axial translation can be defined as Eq. 1. Where $u_t$ is the control input, $m_t$ is mass of the moving part, $c_t$ and $k_t$ are damping and elastic coefficients respectively, while $\ddot{x}$, $\dot{x}$, and $x$ are the motion variables up to second order derivative. Upon actuation, the actual motion response ($v_t$) achieved can be measured to deduce the motion error as given in Eq. 2. The control $u_t$ and $v_t$ are values in the Euclidean space; thus, their difference (control error) is assessed in a similar space. In classical settings, the error value can be controlled with the PID control gains in Eq. 3 such that the $w_t$ is an adapted control input that is issued to ensure $e(t) = 0$ for a $u_t$ value.

$$u_t = m\ddot{x}_t + c\dot{x}_t + kx_t \quad (1)$$

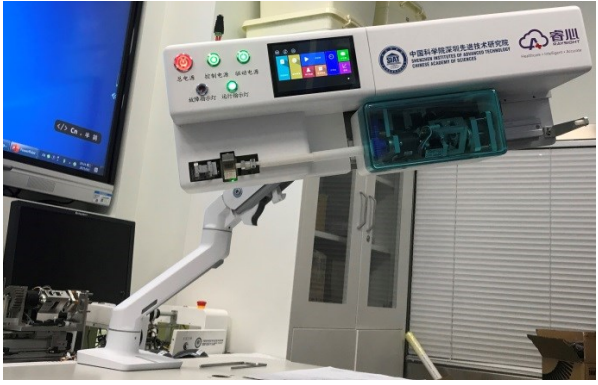

Figure 1. Mechanism of the Slave Device for Robot-assisted PCI

$$e_t = u_t - v_t \quad (2)$$

where $w_t$ is the control input required at time $t$; to model and eliminate the instantaneous error $e(t)$; $k^{j:=\{p,i,d\}}$ are the proportional, integral and derivative gains of the controller, respectively. Effectiveness of this classical PID is dependent on the suitability of the gain tuning. Unlike in the existing studies discussed above [11-18], instead of using either Ziegler-Nichols or supervised tuning methods to determine the appropriate gain values, a deep reinforcement learning is developed for same purpose.

$$w_t = k_t^p e_t + k_t^i \int e_t + k_t^d \frac{e_t}{dt} \quad (3)$$

*i. Reinforcement Learning (RL) Basics:*

In the baseline reinforcement learning, there exists an agent who interacts with its environment and learns to take suitable action needed to achieve a given goal. The action $(k_t^j)$ transits the robot from an old state $(s_t)$ to a new one $(s_{t+1})$ at each time-step, and it either gets rewarded or punished $(r_t)$ based on its executed actions. The basic reinforcement learning is an optimization problem with an infinite-horizon discrete-time Markov decision process (MDP) model given as Eq. 4, where the probabilities $(P_\pi)$ and rewards $(R_\pi)$ are unknown. To solve this problem, we will learn the estimates of optimal values for each action. This is given defined as the expected sum of future rewards if that action is taken and the optimal policy is followed afterwards. For a specified policy, the actual value of an action $k_t^j$ taken at state $s_t$ is defined with Eq. 4. A major task in RL is for the RCS to learn $\left(\pi_*\left(s_t|k_t^j\right)\right)$ an optimal policy that maximizes expected cumulative reward. The chosen action $(k_t^j)$ is completely determined by $\pi$ with a probability of transition from state $s_t$ to $s_{t+1}$ for the given action being $P_\pi(k_t^j = k^j | s_t = s_{t+1})$.

$$Q_\pi(s_t, k_t^j) = \sum \left[\left(R_\pi(s_{t+1}|s_t, k_t^j) + \gamma V(s_t)\right) | P_\pi(s_t, k_t^j, \pi)\right] \quad (4)$$

Where $V(s_t) = \max_k Q^*\left(s_t, k_{t+1}^j\right)$ is the cumulative reward maximized, and $\gamma \in [0,1]$ is a discount factor that is used to decide preference for immediate or future rewards. The optimal policy is derived by chosen optimal values $\left(Q_\pi(s_t, k_t^j) = max_\pi\left(Q_\pi(s_t, k_t^j)\right)\right)$ using highest valued action in each state. This is a typical Q-learning update, and $\theta_t$ is the network parameters. The target is a collection of target $Q$-values of the next states $S_{t+1}^j$ taken by the RCS for all valid actions $K_t^j$. It can be defined as $\hat{V}_t^Q := R_{t+1} + \gamma \max_a Q(S_t, K_t^j; \theta_t)$ ; the optimally parameterized value function $Q(s_t, k_t^j; \theta_t)$ updates the parameters as in Eq. 5 after taking actions in successive time-step for at size of $\alpha$.

$$\theta_{t+1} = \theta_t + \alpha \left(\hat{V}_t^Q - Q(S_t, K_t^j; \theta_t)\right) \nabla_{Q_t} Q(S_t, K_t^j; \theta_t) \quad (5)$$

*ii. Episodic Update with Policy Transfer:*

Estimates of optimal values and policies are established on the Q-learning approach in a form of temporal learning process, and learning update is done based on stochastic gradient descent taken to have the robot learn and transfer policy maps so as to maximize the cumulative reward. Thus, a deep reinforcement learning system is incorporated to learn and transfer policies between different episodes in solving such MDPs. Often representing the deep neural network and related modern soft-computing functions can be developed for learning and adapting the optimal policy $\pi_t(s_t|k_t^j)$. The artificially intelligent algorithms perform specialized learning that well suited the dynamic settings of the robot's environment. In this study, a deep Q-network, a multi-layered model of deep neural network augmented with LSTM is developed for transferring policies between training episodes. The deep Q-network learns from a $\Re^{n\times m}$ state-action space using source and target networks trained on experience replay trained on a vector of given state, action output, next state, and network parameters, *i.e.* Q$(s_t, k_t^j, s_{t+1}; \theta_t)$ Mnih *et.al* [22]. Parameters of the target network are updated from the source network at a given time step interval to stabilize learning and improve performance. Thus, target used by the deep Q-network is defined in Eq. 6 where the parameters of the network are updated after the $\tau$ time steps.

$$\hat{V}_t^{DQ} = R_{t+1} + \gamma \max_{k\in K} Q(S_t, K_{t+1}^j; \hat{\theta}_t^{DQ}) \quad (6)$$

Where $\tau$ is the time step for transferring parameters of source network to update the target network. This update strategy eliminates the chasing-own-tail scenario and the network can achieve a reduced computational complexity. The transition series from an episode can be saved as an experience tuple $\xi := (s_t, k_t^j, r_t, s_{t+1})$ and the corresponding rewards $r(t)$ in a temporary memory for knowledge recall. The target Q-table $\left(\hat{V}_{\pi t}^{DQ}(S_t, K_{t+1}^j; \hat{\theta}_t)\right)$ is updated at every $\tau$ step and it, each time, contains policy transferred from previous episodes that can be used to arrive at states $(S_{t+1})$ when valid actions $(k_t^j \in K_t^j)$ in an action space is taken.

*iii. Deep reinforcement learning PID (DRL-PID):*

The RL agent continuously learns as it interacts with the environment and automatically fixes the PID control gains $k_{t,y|z}^j = \{(k_{t,y}^p, k_{t,y}^i, k_{t,y}^d), (k_{t,z}^p, k_{t,z}^i, k_{t,z}^d)\}$ which are the actions taking by the robot learning the control policy $\pi$. The subscripts $y|z$ indicate the PID is either for axial translation or rotation motion. First, base gains ${}_b k_{t,x}^j$ are chosen as the PID control parameters from a uniformly random distribution such that ${}_{min}k_{t,x}^j \leq {}_b k_{t,x}^j \leq {}_{max}k_{t,x}^j$. This eliminates human bias and needs for specifying rules or data to intelligently tune the classical PIDs. According to the range of gain states, the axial translation displacement $(x)$ and velocity $(\dot{x})$ are defined as equal interval variables needed to derive the reward component ($r_t$) of the DRL, as in Eq. 8. In this sense, $r_t$ at step $t$ can be defined with the Gaussian function in Eq. 8, where the hyperparamter ($\sigma^2$) is used to define the shape of the Gaussian function. The hyperparamter can be carefully chosen to expand the kernel function's space, reward agent without direct knowledge of human, and avoid learning overfitting and underfitting. $x_{req}$ is the actual distance, and $x_t$ is the set-point distance used for higher derivatives in Eq. 1. For a 1-DoF RCS, $r_t$, $s_t$ and $x_t$ are scalar values in the Euclidean space. The immediate and aggregate rewards are obtained as $R_t = \sum_{t=0}^{\infty} \gamma^t r_t$, a reward discounted at rate $\gamma \in [0,1)$. The reward at each step $r_t(a_t, x_t)$ is defined to satisfy motion constraint requirements that guide the axial motion in the RCS. The Gaussian function rewards the agent positively high when $e_t$ in the current state ($a_t$) is close to zero and rewards lower based on the instantaneous error value.

$$R_T(X_T, X_{REQ}) = \frac{1}{2\Pi\Sigma} \text{EXP}\left(-\frac{\|X_T - X_{REQ}\|^2}{2\Sigma^2}\right) \quad (8)$$

III. MODEL IMPLEMENTATION AND ANALYSIS

The deep reinforcement learning model proposed with sample-efficient and episodic policy transfer is validated in this section. First, we present the implementation details and different experiments done to validate the performance of our proposed self-adaptive learning-based model. Next, the validation and performance evaluation results obtained are discussed. Finally, real-time application of the model is demonstrated with our RCS and the results are presented.

*A. Robot Simulation*

The CAD model of the PCI robot in Fig. 1 was designed in Solidworks® (Dassault Systems Solidworks Corp., USA) and exported via unified robot description format (URDF) into CoppeliaSim (Coppelia Robotics, Switzerland) for real-time simulation. Parts controlling the axial translation and rotation were imported separately and assembled in the CoppeliaSim environment, while their static and dynamics variables were parameterized for apposite emulation of PCI catheterization in a CathLab. To enable interaction of the agent in the simulation environment, robot navigation were defined on prismatic and revolute joints added for distinct control, while sensor components were added for motion feedback data in CoppeliaSim. Lua scripts were written to communicate and manipulate the robot's components from external Python implementation, and cross-language scripts interactions were achieved via threaded communication in a proprietary dynamic link library of CoppeliaSim.

*B. Agent Episodic Learning Model*

The robot agent was trained with a deep learning model that consists of a double source network (actors) and single target network (learner) to adaptively approximate the Q-learning function. The first actor is used to make the actual predictions on what action to take, while, since PID gains could vary greatly depending on the agent's environment and state, the other actor is used to bound the PID gains $\left(k_t^p, k_t^i, k_t^d\right)$ into a specified range such that each gain is greater and lower than specified minimum and maximum values, respectively. Gain tuning (action selection) limits is achieved by stopping growth of the action in a direction that is reaching the boundary set by the actor network. In contrary, the learner network is used to intuitively track the experiential action desired for the agent to take at a given state and environment factors. Structures of the networks are given in Fig. 2a. Generally, each network consists of four dense layers whereas, unlike in the bounding actor network, an adaptive moment optimization layer is added in the predicting actor and learner networks. A hyperbolic tangent function $(tanh(\cdot))$ is used for value limiting in the bounding network. This comparative performs better than adding loss function proposed for deep deterministic policy gradient algorithm [23].

The Q-network is trained off-policy with memory replay in target network to pass experience between episodes; and a fixed discount rate ($\gamma$) is added for online training to avoid saturation of the policy network and allay collision while exploiting stochastically with MDP. During training, the agent is attentive to improve its policy and only actions in a bounded region of the actions space are executed while actions with higher rewards are explored over time. The reward policy is specified for each action as defined in Eq. 8 where the kernel space is determined with Eq. 9.

$$\sigma\left(s_t, k_t^{j\forall\{p,i,d\}}\right) = \begin{cases} x_{req} - x_t < \mathcal{E} & 1.0 \\ x_{req} - x_t > 1.5 * \mathcal{E} & -1.5 \\ x_{req} - x_t > 10 * \mathcal{E} & -5.0 \\ x_{req} - x_t < \mathcal{E} \wedge fastConv & +5.0 \\ else & -1.0 \end{cases} \quad (9)$$

Where $\mathcal{E}$ is the target minimum axial error and $fastConv$ is Boolean values that decides if the agent learns and converges faster in a given step (proxy on 80% of time allocated for PID gain tuning).

*C. Model Implementation*

The deep reinforcement model was implemented with Tensorflow Keras framework on a desktop computer with Intel® Core i7 processor (3.4 GHz each), 48 GB RAM and Nvidia GTX 1080 graphics card. The network training was done on $Q_0$ for each time step with a mini-batch (size = 32) state-action-reward experience tuple $\left(\xi := \left(s_t, k_t^j, r_t, s_{t+1}\right)\right)$ randomly sampled into buffer for stochastic experience replay. The replay buffer is bounded to only store most recent state-action transitions while older knowledge are removed to utilize rewarding experience for transitions and also to avoid the chasing-own-tail scenario. The exploration rate, set as 0.005, with linearly piecewise decaying factor, set as 1 to 0.01, was applied; while the discount factor was set as 0.85. The baseline and bounding parameters used for gain tuning are [$k_o^p$ = 1.2, $k_o^i$ = 1.0, $k_o^d$ = 0.01] and [1e-2, 1.0], respectively. The D-term of the controller is further scaled with a factor of 0.1 to increase stability in simulation and experimental applications. To ensure proper gain responsiveness to error values and mitigate the degree to which the DRL model overshoots a setpoint, we initialize the setpoint to 0.0 in the first 10 tuning turns of each iterative learning step.

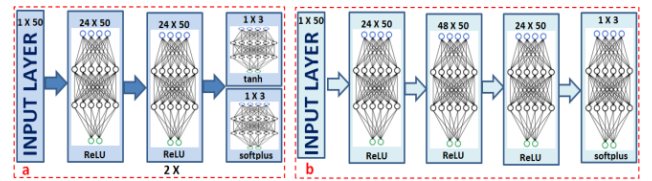

Figure 2.   Mechanism of the Slave Device for Robot-assisted PCI

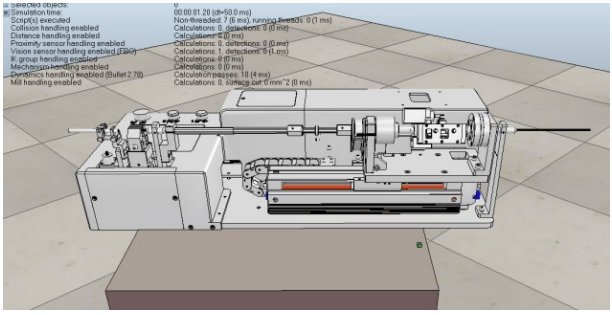

Figure 3. Simulation of the Slave Robot in CoppeliaSim

*D. Implementation Results*

The proposed DRL-PID algorithm was applied for the 1-DoF RCS in Fig. 1 in a simulation trial performed with the CoppeliaSim simulator as shown in Fig. 3. In this study, we manipulated the axial motion to obtain training data for the DRL agent. The learning process converged in 20 episodes with an intra-episodic learning iteration of 100 step units. An intra-episodic training terminates based on Eq. 9 and the reward is set accordingly. The learning process and rewards aggregated during each episode are presented in Fig. 4. The training steps per each episode in Fig. 4a shows the agent could acquire and build knowledge of the environment over time while the average feedback (*i.e.* the states) observed during the tuning process at this step is presented in Fig. 4c. This shows how the agent tunes the control gains to stability. The total reward per episode in Fig. 4b shows that the agent learns over time and transfers its experience across episodes. This is further is explained in Fig. 4d which, with effects of Eq. 8, shows the mean intra-episode reward aggregation. Also, the reward policies obtained in each intra-episodic step (Fig. 4d) also inform that the agent built knowledge iteratively from response to given set-points, and the experience gained are used to set appropriate gains of the controller within each episode.

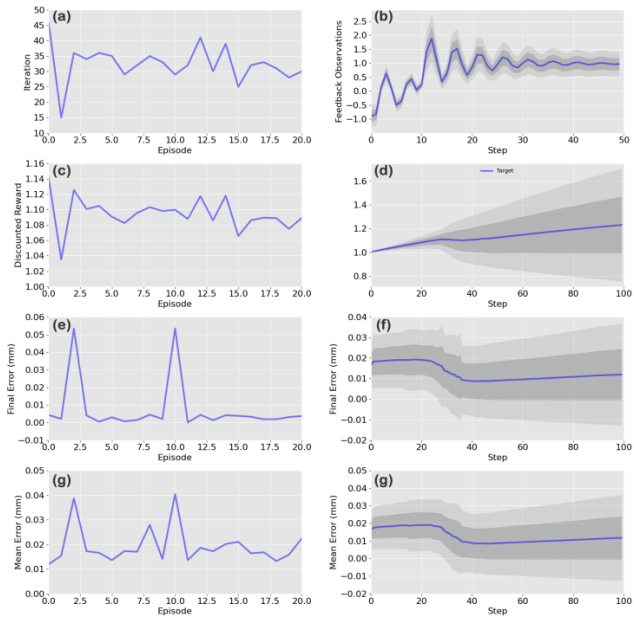

Figure 4. Simulation of the Slave Robot in CoppeliaSim: (a) total steps per episode; (b) intra-step feedback obtained in last step of training; (c) discounted rewards per episode; (d) mean total reward in the step; (e) final error per episode and mean of last ten error values per step; (f) final error per episode and mean of last ten error values per step.

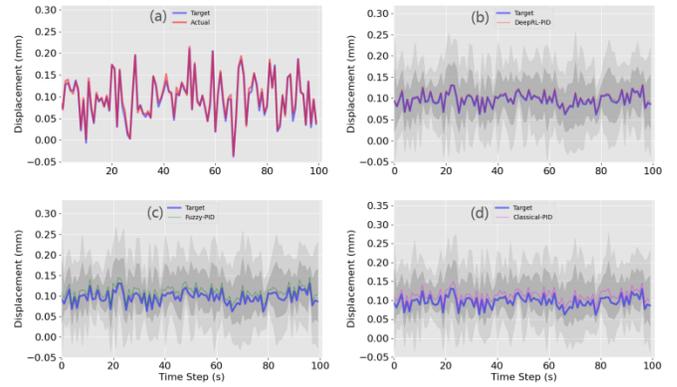

Figure 5. PID-based set-point axial navigation (a) result from DRL-PID tuning agent in a trial. Average results from (b) DRL-PID; (c) Fuzzy-PID, and (d) Classical-PID for multiple random trials.

To validate this, the final and mean final error per episode along with the mean of last ten final error values and their average from all steps per episode are computed as presented in Fig. 4e-g, respectively. These validates that the agent has overall minimal mean errors with much lower final errors. Thus, the agent's knowledge improves better during each of the intra-episodic learning.

IV. VALIDATION AND PERFORMANCE EVALUATION

*A. Simulation Study: Model Validation*

First, the model was applied for motion control of the slave-side RCS in CoppeliaSim. The agent was applied to aid navigation control of the RCS by providing apt control gains needed for randomly generated displacement values. Using a normal distribution with mean and standard deviation of 0.1, and 0.05, respectively, the random values were set as input to trigger axial displacement of the RCS as shown in attached movie (S1), while the actual distance moved by the robot were recorded with in-built sensors in CoppeliaSim. The results obtained within the simulation environment are displayed in Fig. 5a. With the proposed DRL-PID, the system is capable of accurate and responsive tracking which is a key process for tool navigation in teleoperated robotic catheterization. The maximum navigation and RMS errors in the figures are 0.004 and 0.007 mm, respectively. In this regards, it could be understood that the agent adaptively moves with the states to appropriately tune the PID gains (*i.e.* select apt actions) based on the required conditions.

*B. Simulation Study: Performance Evaluation*

Performance of the proposed DRL-PID was compared with conventional tuning methods which are based on was classical, and adaptive fuzzy tuning. The approaches were implemented in python language and used to govern axial navigation of the RCS in CoppeliaSim. For classical method, the PID control gains were obtained using Ziegler-Nichols system, while triangular membership functions with 49 *if- else* rules are used in the fuzzy approach. The rules are set based on the error term and its first derivative taken as inputs applied to decide the gain values, as used in previous studies [14, 19].

To evaluate the performances of the tuning methods, 100 set-point samples were drawn from a Gaussian distribution with similar mean and standard deviation used above. The values were set as instantaneous axial displacement of the RCS in during 10 different simulation trials using each tuning method, while the actual distance moved by the axial motor were recorded from the coordinate sensor in the simulated robot. As shown in Fig. 6b-d, the deep reinforcement agent caused a mean and RMS errors of $0.003 \pm 0.0058$ mm compared to $0.012 \pm 0.009$ mm and $0.015 \pm 0.011$ mm observed in the classical and fuzzy-based methods, respectively. Thus, it can be understood that the deep reinforcement learning agent performed a better tuning that as it offered the most stable set-point in the navigation control of the RCS, it is also established that it offers a more adaptive and self-tuning system of the PID which can be applied in the presence of communication issues where both classical and conventional methods fails.

*C. Experimental Study*

Similarly, the model was applied in an environment with the actual slave-side device of our RCS shown in Fig. 6. All components in the robot and their functionalities are found in our previous study [21]. In this current work, we focus on tracking the displacement commands received from control knob of the master interface (which is integrated potentiometer that provides displacement values). Due to hardware constraints in microprocessor used for the RCS (8GB Raspberry Pi-4 ARM), we exported the final tuning parameters and applied in the robotic system. For operation consistency, we only set a constant value of 1 mm, which was used to train the agent, as displacement command for the real time motion control in the RCS. The operation of the robot while executing the motion is recorded as movie (S2). Also, Fig. 7 shows the error values obtained while the RCS executed the control command iteratively.

## V. CONCLUSION AND FUTURE WORKS

In this study, a first time approach of deep reinforcement learning for motion control during robotic catheterization procedures is developed. The model is based on consists of a double source and single target actor-learner networks that approximate Q-function. A deep learning mode is integrated for episodic policy learning and transfer during the reinforcement learning to adaptively tune PID gains for tool navigation purposes. Validated of the trained reinforcement model was applied for axial navigation of a RCS model and offered satisfactory results. More validation studies with *in-silico* and *in-vitro* experimental settings are still needed. Also, the DRL agent was trained several times but in short episodes. This can be attributed to the PID base gains values, the hyper parameters, and are reward policy function applied for training. The latter has a kernel space obtained with the Gaussian function (Eq. 8); thus, more validations are needed.

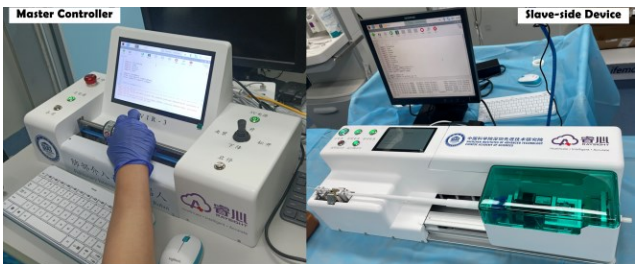

Figure 6. Master-slave Robotic Catheter System for PCI

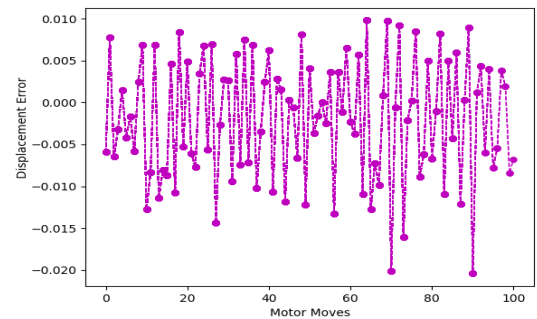

Figure 7. Simulation of the Slave Robot in CoppeliaSim